\documentclass[conference]{IEEEtran}

\usepackage{lineno,hyperref,amsmath}
\usepackage{float}
\usepackage{graphicx}
\usepackage{lscape}
\usepackage{booktabs}
\usepackage{subcaption}
\usepackage{multirow}

\modulolinenumbers[5]

\begin{document}
\title{Nonparallel Hyperplane-based Classifiers for Multicategory Classification}


\author{\IEEEauthorblockN{Reshma Khemchandani, Pooja Saigal}
\IEEEauthorblockA{Department of Computer Science,\\ Faculty of Mathematics and Computer Science, \\ South Asian University,\\ Delhi, India\\
Email: reshma.khemchandani@sau.ac.in\\Email: pooja.saigal@students.sau.ac.in}}

\maketitle

\begin{abstract}
	Support vector machine (SVM) is widely used for solving classification and regression problems. Recently, various nonparallel hyperplanes classification algorithms (NHCAs) have been proposed, which have comparable classification accuracy as SVM but are computationally more efficient. All these NHCAs are originally proposed for binary classification problems. Since, most of the real world classification problems deal with multiple classes, these algorithms are extended in multicategory scenario. In this paper, we present a comparative study of four NHCA i.e. Twin SVM (TWSVM), Generalized eigenvalue proximal SVM (GEPSVM), Regularized GEPSVM (RegGEPSVM) and Improved GEPSVM (IGEPSVM) for multicategory classification. The multicategory classification algorithms for NHCA classifiers are implemented using One-Against-All (OAA), binary tree-based (BT) and ternary decision structure (TDS) approaches and the experiments are performed with benchmark UCI datasets. 
\end{abstract}

\IEEEpeerreviewmaketitle

\section{Introduction}
Support vector machines (SVMs) \cite{cortes1995support},\cite{burges1998tutorial} have been widely used for classification and regression problems. SVM involves the minimization of a convex quadratic function, subject to linear inequality constraints and thus generates a hyperplane that separates the two classes. In contrast to SVM, Mangasarian et al. proposed proximal support vector machine classifiers \cite{mangasarian2001proximal} that determines two parallel planes such that each plane is proximal to one of the two classes to be classified and as far as possible from the other class. The binary classification problem can also be formulated as a generalized eigenvalue problem (GEP) \cite{parlett1980symmetric}, as proposed by Mangasarian et al. and is termed as GEPSVM \cite{mangasarian2006multisurface}. This formulation for GEPSVM differs from that of SVM; since, instead of generating one hyperplane that separates the two classes, it determines two hyperplanes that approximate the two classes. Also, SVM solves a quadratic programming problem (QPP) and GEPSVM solves two GEPs. Therefore, GEPSVM is faster than SVM. In the past few years, various modifications for GEPSVM have been proposed like Regularized GEPSVM (RegGEPSVM) \cite{guarracino2007classification} and Improved GEPSVM (IGEPSVM) \cite{shao2013improved}. On the lines of GEPSVM, Jayadeva et al. proposed TWSVM \cite{khemchandani2007twin} which is a nonparallel plane classifier for binary data classification. TWSVM solves two smaller-sized QPPs and generates two nonparallel hyperplanes such that each is closer to one of the two classes and is as far as possible from the other.

Most of the SVM-based classifiers cater to binary classification problems, but real world problems deal with multiple classes. Researchers have been trying to extend these classifiers to multicategory scenario. The two most popular approaches for multiclass SVMs are One-Against-All (OAA) and One-Against-One (OAO) support vector machines \cite{hsu2002comparison}. OAA-SVM implements a series of binary classifiers where each classifier separates one class from rest of the classes, but it could lead to class imbalance problem, due to huge difference in the number of samples. For a K-class classification problem, OAA-SVM requires $(K-1)$ SVM classifiers. In case of OAO-SVM, the binary SVM classifiers are determined using a pair of classes at a time. Thus, it formulates upto $(K*(K-1))/2$ binary SVM classifiers and increases the computational complexity. Lei et al. propose Half-Against-Half (HAH) multiclass-SVM \cite{lei2005half}. HAH is built via recursively dividing the training dataset of K classes into two subsets of classes. Shao et al. propose a decision tree twin support vector machine (DTTSVM) for multi-class classification \cite{shao2013best}, by constructing a binary tree based on the best separating principle, which maximizes the distance between the classes. Khemchandani et al. proposed ternary decision structure (TDS) \cite{khemchandani2015color} for multicategory classification for TWSVM.

In this paper, we present a comparative study of four NHCAs i.e. TWSVM, GEPSVM, RegGEPSVM and IGEPSVM in multicategory framework. We explore three approaches for multicategory extension, namely OAA, BT and TDS. In case of BT, the data is recursively divided into two halves and a binary tree of classifiers is created. BT determines $(K-1)$ classifiers for a K-class problem. It is observed that tree-based approaches (BT, TDS) are computationally more efficient than OAA, in learning the classifier. The TDS approach outperforms the other two, in terms of classification accuracy. 

The paper is organized as follows: Section ~\ref{sec:gepsvm} gives a brief introduction of NHCA algorithms. Section ~\ref{sec:prop} describes the proposed work which is followed by experimental results in Section ~\ref{sec:exp}. Finally, the paper is concluded in Section ~\ref{sec:con}.

\section{NONparallel hyperplanes classifiers}
\label{sec:gepsvm}
NHCAs determine a hyperplane for each class, such that each hyperplane is proximal to the data points of one class and at maximum possible distance from the data points of the other class. In this section, we briefly outline the linear version of GEPSVM-based and TWSVM classifiers.

\subsection{GEPSVM}
GEPSVM \cite{mangasarian2006multisurface} generates two nonparallel hyperplanes by solving two GEPs of the form $Gz=\mu Hz$, where G and H are symmetric positive semidefinite matrices. The eigenvector corresponding to the smallest eigenvalue of each GEP determines the hyperplane. 
The data points belonging to classes +1 and -1 (referred as positive and negative classes) are represented by matrices A and B, respectively. Let the number of patterns in classes 1 and -1 be given by $m_1$ and $m_2$, respectively. Therefore, the size of matrices A and B are $(m_1 \times n)$ and $(m_2 \times n)$. The GEPSVM classifier determines two nonparallel planes
\begin{eqnarray}
x^Tw_1+b_1=0 \mbox{  and  } x^Tw_2+b_2=0, \label{planes}
\end{eqnarray} 
so as to minimize the Euclidean distance of the planes from the data points of classes 1 and -1, respectively. This leads to the following optimization problem:
\begin{eqnarray}
\underset{w,b \neq 0}{Min} ~~\frac{\lVert Aw+eb \rVert ^2/\lVert [w,b]^T \rVert ^2}{\lVert Bw+eb \rVert ^2/\lVert [w,b]^T \rVert ^2}, \label{gep}
\end{eqnarray} 
where $e$ is a vector of ones of appropriate dimension and $\lVert \cdot \rVert$ denotes the $L_2$ norm. It is implicitly assumed that $(w,b) \neq 0 \Rightarrow Bw+eb \neq 0$ \cite{mangasarian2006multisurface}. 
The optimization problem (\ref{gep}) is simplified and regularized by introducing a Tikhonov regularization term \cite{tikhonov1977solutions} as follows:
\begin{eqnarray}
\underset{w,b \neq 0}{Min} ~~ \frac{(\lVert Aw+eb \rVert ^2 + \delta \lVert [w,b]^T \rVert ^2 )}{\lVert Bw+eb \rVert ^2} \label{gep2},
\end{eqnarray}
where $\delta > 0$ is the regularization parameter. This, in turn, leads to the Rayleigh Quotient of the form
\begin{eqnarray}
\underset{w,b \neq 0}{Min} ~~\frac{z^Gz}{z^THz} \label{gep3},
\end{eqnarray}
where G and H are symmetric matrices in $R^{(n+1)\times (n+1)}$ defined as
\begin{eqnarray}
&&G=[A ~~ e]^T \times [A ~~ e]+ \delta \times I \mbox{ for some }\delta>0, \nonumber \\
&&H=[B ~~ e]^T \times [B ~~ e], \mbox{  and  } z=[w,b]^T.
\end{eqnarray}
$I$ is an identity matrix of appropriate dimensions. Using the properties of the Rayleigh Quotient \cite{mangasarian2006multisurface},\cite{parlett1980symmetric}, the solution of (\ref{gep3}) is obtained by solving the GEP
\begin{equation}
Gz=\mu Hz, ~~ z\neq 0, \label{gep4}
\end{equation}
where the global minimum of (\ref{gep3}) is achieved at an eigenvector
corresponding to the smallest eigenvalue $\mu_{min}$ of (\ref{gep4}). Therefore, if $z_1$ denotes the eigenvector corresponding to $\mu_{min}$, then $[w_1,b_1]^T=z_1$ determines the plane $x^Tw_1+b_1=0$ that is close to the positive class. Next, we define another minimization problem analogous to (\ref{gep}) by interchanging the roles of A and B. The eigenvector $z_2$ corresponding to the smallest eigenvalue of the
second GEP yields the plane $x^Tw_2+b_2=0$, which is close to points of class -1.

\subsection{RegGEPSVM}
Guarracino et al. \cite{guarracino2007classification} modified the formulation of GEPSVM, so that a single GEP can be used to generate both the hyperplanes. The GEP $Gz=\mu Hz$ is transformed as $G^*z=\mu H^*z$ where
\begin{equation}
G^*=\tau_1 G-\delta_1H, ~~ H^*=\tau_2 H-\delta_2G.
\end{equation}
The parameters $\tau_1$, $\tau_2$, $\delta_1$ and $\delta_2$ are selected, such that the $2 \times 2$ matrix
\begin{equation}
\Omega=\begin{bmatrix}
\tau_2 & \delta_1 \\
\delta_2 & \tau_1
\end{bmatrix}
\end{equation} 
is nonsingular. The problem $G^*z=\mu H^*z$ would generate same eigenvectors as that of  $Gz=\mu Hz$. An associated eigenvalue $\lambda^*$ of the transformed problem is related to an eigenvalue $\lambda$ of the original problem by
\begin{equation}
\lambda=\frac{\tau_2\lambda^*+\delta_1}{\tau_1+\delta_2 \lambda^*}.
\end{equation}
By taking $\tau_1=\tau_2=1$ and $\nu_1=-\delta_1$, $\nu_2=-\delta_2$, the problem becomes
\begin{eqnarray}
\underset{w,b \neq 0}{Min} ~~\frac{\lVert Aw+eb \rVert ^2+ \nu_1\lVert Bw+eb \rVert ^2}{\lVert Bw+eb \rVert ^2+ \nu_2 \lVert Aw+eb \rVert ^2}, \label{reg}
\end{eqnarray} 
When $\nu_1,~\nu_2$ are non negative, $\Omega$ is non-degenerate, then the eigenvectors related to the minimum and maximum
eigenvalue of (\ref{reg}) would be same as obtained by solving the two GEPSVM problems.

\subsection{IGEPSVM}
IGEPSVM \cite{shao2013improved} replaced the generalized eigenvalue decomposition by standard eigenvalue decomposition which resulted in solving two simpler optimization
problems and introduced a parameter to improve the generalization ability. IGEPSVM formulated the two problems as
\begin{eqnarray}
\underset{w,b \neq 0}{Min} ~~\frac{\lVert Aw+eb \rVert ^2}{\lVert w \rVert ^2+ b^2} -\nu\frac{\lVert Bw+eb \rVert ^2}{\lVert w \rVert ^2+ b^2}, \label{igep}
\end{eqnarray} 
where $\nu>0$ trade-off factor between the two terms in the objective functions. Thus, IGEPSVM has a bias factor for different classes and is particularly useful for solving the imbalance dataset problems. By introducing a Tikhonov regularization term and solving its Lagrange function by setting partial derivatives concerning the primal variable equal to zero, we get
\begin{equation}
((M^T+\delta I)-\nu H^T)z=\lambda z,
\end{equation}
where $M=[A~~e]^T[A~~e], ~H=[B~~e]^T[B~~e]$, $z=[w,b]^T$ and $\lambda$ is Lagrange multiplier. $I$ is an identity matrix of appropriate dimensions. The second problem can be defined analogous to (\ref{igep}) by interchanging A and B, as discussed for GEPSVM.

The three GEPSVM-based classifiers can be extended to nonlinear classifiers by considering the following kernel-generated surfaces instead of hyperplanes.
\begin{eqnarray}
K(x^T,C^T)u_1+b_1=0 ,\nonumber \\
 K(x^T,C^T)u_2+b_2=0, \label{kplanes}
\end{eqnarray}
where $C^T=[A~~B]^T$ and $K$ is an appropriately chosen kernel.

\subsection{Twin Support Vector Machine}
TWSVM \cite{khemchandani2008mathematical} is a binary classifier that determines two nonparallel planes by solving two smaller-sized QPPs such that all patterns do not appear in the constraints of either problem at the same time. The TWSVM classifier is obtained by solving the following QPPs where $e$ is a vector of ones of appropriate dimensions. $c > 0$ is trade off factor between error variable $q$ due to class $-1$ and distance of hyperplane from its own class $1$.
\begin{eqnarray}
\min_{w,b,q} & \frac{1}{2}(Aw+eb)^T (Aw+eb)+ce^Tq	\nonumber\\
{\mbox {subject to}} &  -(Bw+eb)+q \geq e,~~	q \geq 0 \label{TWSVM1}
\end{eqnarray}
The Wolfe dual \cite{mangasarian1993nonlinear} of (\ref{TWSVM1}) is as follows:
\begin{eqnarray}
\max_\alpha &  e^T\alpha -\frac{1}{2} \alpha^TG(H^TH)^{-1}G^T \alpha \nonumber\\
{\mbox {subject to}} & 0\leq \alpha \leq c. \label{DTWSVM1} 
\end{eqnarray}
Here, $H =[A \; e]$, $G=[B \;e]$, and the augmented vector $u=[w,b]^T$ is are given by
\begin{equation}
u=-(H^TH)^{-1}G^T\alpha.
\end{equation}
Here, $\alpha=(\alpha_1,\alpha_2,...,\alpha_{m2})^T$ are Lagrange multipliers. The second hyperplane can be obtained by interchanging A and B in (\ref{TWSVM1}). The test patterns are classified according to which hyperplane a given point is closest to.

\section{Comparison of NHCAs for multicategory classification}
\label{sec:prop}
Most of the SVM-based classifiers are originally designed for binary classification. However, to use these classifiers in real life situations, they must be extended in multicategory framework. Two most common approaches are: directly considering all data in one optimization formulation, while the other is by constructing and combining several binary classifiers \cite{hsu2002comparison}. The first option considers all classes at one time and leads to a very complex and computationally expensive optimization problem. Whereas, the second approach builds a number of smaller-sized classifiers and is computationally more efficient. In our work, we propose extension of NHCAs using OAA, BT and TDS approaches for multicategory classification. 

\subsection{Extending NHCA classifiers using One-against-all (OAA) approach}
In order to solve a K-class classification problem using OAA multicategory approach, we construct $(K-1)$ binary NHCA classifiers. Here, each classifier represents a pair of nonparallel hyperplanes. The $i^{th}$ classifier $(i=1~to~K)$ is trained with all the patterns in
the $i^{th}$ class with positive labels and all other patterns with
negative labels. With $m$ data patterns $((x_j,y_j),~j=1~to~m)$, the matrices $A=\{x_p:~y_p=i\}$ and $B=\{x_q:~y_q \neq i\}$ are created. The patterns of $A$ and $B$ are assigned labels $+1$ and $-1$ respectively. This data is used as input for GEPSVM in (\ref{gep2}), RegGEPSVM in (\ref{reg}), IGEPSVM in (\ref{igep}) and TWSVM in(\ref{DTWSVM1}) to generate $K$ classifiers. To test a new pattern, we find its distance from all the $K$ hyperplanes corresponding to positive classes and the actual class label of nearest hyperplane is assigned to the test pattern i.e. the test pattern $x\in R ^n$ is assigned to class $r(r=1~to~K)$, depending on which of the $K$ hyperplanes given by (\ref{planes}) it lies closer to, i.e.
\begin{eqnarray}
x^Tw^{(r)}+b^{(r)}= \min_{l=1:K} \frac{\vert x^Tw^{(l)}+b^{(l)}\vert}{\Vert w^{(l)} \Vert_2 }, \label{Tminplane}
\end{eqnarray}
where $\vert .\vert$ is the absolute distance of point $x$ from the plane $x^Tw^{(l)}+b^{(l)}=0$.

\subsection{Extending NHCA classifiers through Binary Tree-based (BT) approach}
BT builds the classifier model by recursively dividing the training data into two groups and creates a binary tree of classifiers \cite{khemchandani2015color}. At each level of the binary tree, training data is partitioned into  two groups by applying k-means (k=2) clustering \cite{hartigan1979algorithm} and the hyperplanes are determined for the two groups using GEPSVM-based classifiers; use (\ref{gep2}) for GEPSVM, (\ref{reg}) for RegGEPSVM, (\ref{igep}) for IGEPSVM and (\ref{DTWSVM1}) for TWSVM. This process is repeated until further partitioning is not possible. The BT classifier model thus obtained can be used to assign the label to the test pattern. The distance of the new pattern is calculated form both the hyperplanes, at each level and the group with nearer hyperplane is selected, as given is (\ref{Tminplane}). Repeat it, till a leaf node is reached and assign the label of leaf node to the test pattern. TB determines $(K-1)$ NHCA classifiers for a K-class problem, but the size of the problem diminishes as we traverse down the binary tree. For testing, TB requires at most $\lceil log_2K \rceil$ binary evaluations.

\subsection{Extending NHCA classifiers through Ternary Decision Structure (TDS)}
TDS evaluates all the training points into an `i-versus-j-versus-rest' structure. During the training phase, TDS recursively divides the training data into three groups by applying k-means (k=2) clustering \cite{hartigan1979algorithm} and creates a ternary decision structure of classifiers. The training set is first partitioned into two clusters which leads to identification of two focused groups of classes and an ambiguous group of classes. The focused class is one where most of the samples belong to a single cluster whereas the samples of an ambiguous group are scattered in both the clusters. Therefore, TDS assigns ternary outputs $(+1, 0, -1 )$ to the samples. TDS partitions each node of the decision structure into at most three groups. 
The cluster labels $(+1, 0, -1)$ are assigned to training data and three hyperplanes are determined using one-against-all approach. This in turn creates a decision structure with height $\lceil log_3K \rceil$. 



In order to extend the capability of NHCA classifiers to handle multiclass data, we propose their use in TDS framework. The training data is partitioned into three classes using k-means clustering 
and the hyperplanes are determined for these groups using NHCA classifiers in OAA approach. To find the three hyperplanes, use (\ref{gep2}) for GEPSVM, (\ref{reg}) for RegGEPSVM, (\ref{igep}) for IGEPSVM and (\ref{DTWSVM1}) for TWSVM. Recursively partition the data sets and obtain classifiers until further partitioning is not possible. Once we have built the classifier model, we can test a new pattern by selecting the nearest hyperplane at each level of the decision node, as in (\ref{Tminplane}) and traverse through the decision structure until we reach a terminating node. The class label of terminating node is then assigned to the test pattern. With a balanced ternary structure, a K-class problem would require only $\lceil log_3K\rceil$ tests. Also, at each level, the number of training samples used by TDS diminishes with the expansion of decision structure. Hence the order of QPP reduces as the height of the structure increases.

\section{Experiments}
\label{sec:exp}
To compare the four NHCAs i.e. GEPSVM, RegGEPSVM, IGEPSVM and TWSVM, we implemented them in multicategory framework with OAA, BT and TDS approaches. The experiments are performed in MATLAB version 8.0 under Microsoft Windows environment on a machine with 3.40 GHz CPU and 16 GB RAM. The simulations are performed with ten benchmark UCI datasets \cite{blake1998uci} and the performance of these algorithms is measured in terms of classification accuracy and computational efficiency in learning the model. The experiments are conducted with 5-fold cross validation and “Accuracy” is defined as follows.
\begin{equation}
Accuracy = \frac{TP+TN}{TP+FP+TN+FN},
\end{equation}
where TP,
TN, FP, and FN are the number of true positive, true negative, false positive and false negative respectively. Classification accuracy of each of the aforementioned methods is measured by the standard five-fold cross-validation methodology \cite{duda2012pattern}. The selected UCI datasets are Iris, Seeds, Dermatology, Wine, Zoo, Ecoli, Glass, Page blocks, Multiple Features and Optical Recognition of Handwritten Digits. Dermatology is referred as Derm, Pageblocks as PB, Multiple Features as MF and Optical Recognition of Handwritten Digits as OD. PB, MF and OD are large datasets, with high number of data samples and features. The grid search method \cite{lanckriet2004learning} is adopted to tune the respective parameters of the four NHCAs and the validation set consists of $10\%$ randomly selected samples from the datasets.

Table \ref{tab:ucilin} shows the classification results of the four NHCAs with three multicategory approaches on ten UCI datasets. The table lists the datasets along with their dimension as $m\times n \times K$, where $m,~ n,~ K$ are the number of data samples, features and classes respectively. For each multicategory classifier, we have reported classification accuracy (Acc in $\%$) along with standard deviation (SD) across the five folds. The best result is shown in bold face. The table also shows the learning time (in seconds) for each of these algorithms. From Table \ref{tab:ucilin}, it is evident that the linear TDS-TWSVM outperforms the other multicategory classifiers in terms of classification accuracy and achieves $89.73\%$ accuracy over the 10 UCI datasets. `Win-Loss-Tie' (W-L-T) ratio gives a count of wins, losses and ties for an algorithm in comparison to other algorithms. From Table \ref{tab:ucilin}, W-L-T for TWSVM and GEPSVM-based classifiers are 8-2-0 and 2-8-0, for classification accuracy. This shows that TWSVM outperforms GEPSVM-based classifiers. Also, W-L-T for OAA, BT and TDS are 1-9-0, 2-7-1 and 6-3-1, which demonstrates that TDS excels other two approaches in terms of classification accuracy. It is also observed that tree-based approaches (BT and TDS) are more efficient than OAA in learning the classifier. BT-RegGEPSVM takes the minimum learning time (1.52 sec), computed as average over 10 datasets. Further, GEPSVM-based classifiers are faster than TWSVM. 

\begin{table*}[t]
	\centering
	
	\begin{tabular}{|l|c|c|c|c|c|c|c|c|c|c|c|c|}
		\hline
		\multirow{5}{*}{ } &
		\multicolumn{3}{c|}{IGEPSVM } &
		\multicolumn{3}{c|}{GEPSVM } &
		\multicolumn{3}{c|}{Reg GEPSVM} &
		\multicolumn{3}{c|}{TWSVM} \\
		\hline
		& OAA & BT & TDS & OAA & BT & TDS & OAA & BT & TDS & OAA & BT & TDS \\
		\hline
		
		& Acc  & Acc  & Acc  & Acc& Acc  & Acc& Acc  & Acc& Acc  & Acc& Acc  & Acc \\
		DATA & SD & SD & SD & SD & SD & SD & SD & SD & SD & SD & SD & SD \\
		SETS & Time & Time & Time & Time & Time & Time & Time & Time & Time & Time & Time & Time \\
		\hline
		\hline
		Iris & 96.00 & 90.00 & 89.33& 96.67 & 95.33 & 96.00 & 96.67 & 95.33 & 96.00	& 95.33 & \textbf{97.33} & \textbf{97.33}\\
		150 $\times$ 4 $\times$	3 & 5.96 &	3.33 &	6.41 &	3.33	& 2.98 &	2.79 &	3.33 &	2.98 &	2.79& 	3.80 &	1.49 &	1.49\\
		& 3.3333 &	\textbf{0.0006} &	0.0013 &	3.2055 &	0.0007 &	0.0011 &	2.3570 &	\textbf{0.0006} &	0.0007 &	4.9441 &	0.0996 &	0.2021\\
		\hline
		
		Seeds & 80.89 &	89.05 &	88.57&	92.38&	93.33&	\textbf{94.29}&	92.38&	93.33&	\textbf{94.29}&	81.42&	93.80&	92.38 \\
		210	$\times$ 7 $\times$ 3	& 3.21&	2.71&	1.99&	3.53&	1.99&	2.13&	3.53&	1.99&	2.13&	4.87&	3.61&	4.57\\
		& 1.9920&	0.0006&	0.0068&	2.0137&	0.0006&	0.0069&	1.7546&	\textbf{0.0004}&	0.0051&	2.3810&	0.0983&	0.2971	\\		\hline
		
		Derm & 88.19&	89.77&	89.75&	84.42&	84.37&	86.32&	84.42&	86.62&	86.32&	92.38&	92.38&	\textbf{95.08}\\
		366	$\times$ 34 $\times$ 6	& 5.15&	4.36&	4.33&	5.88&	4.45&	5.54&	5.88&	4.13&	5.54&	4.57&	4.57&	3.90\\
    	& 8.0134&	0.0059&	0.0078&	3.1642&	0.0051&	0.0072&	5.3190&	\textbf{0.0037}&	0.0108&	4.9702&	0.3337&	0.5164\\		\hline
		
		Wine & 85.52&	92.70&	96.59&	84.83&	93.32&	93.87&	84.42&	94.35&	94.43&	92.17&	94.96&\textbf{	97.17}\\
		178 $\times$ 13 $\times$	3 &	5.95&	4.21&	3.73&	5.20&	6.68&	9.08&	5.88&	6.33&	7.85&	4.07&	3.09&	2.02\\
		&	0.0356&	0.0007&	0.0082&	0.0218&	0.0006& 0.0068&	0.0150&\textbf{	0.0004}&	0.0054&	0.3767&	0.0790&	0.3056\\		\hline
		
		Zoo & 85.10	& 86.10&	87.10&	87.10&	85.05&	87.05&	89.10&	92.05&	93.05&	89.14&	\textbf{94.05}&	93.04\\
		101$\times$ 16 $\times$ 7 	&	4.69&	9.00&	9.79&	6.76&	10.06&	6.78&	5.50&	2.81&	2.78&	6.41&	6.52&	5.72\\
		&	0.0346&	0.0027& 0.0197&	0.0286&	0.0022&	0.0172&	0.0203&	\textbf{0.0018}&	0.0113&	1.0799&	0.1918&	0.2993\\		\hline
		
		Ecoli & 80.25&	83.45&	82.63&	74.14&	81.52&	81.11&	74.14&	80.21&	82.32&	76.14&	82.88&	\textbf{84.42}\\
		327 $\times$ 7 $\times$ 5	&	2.19&	4.02&	2.49&	4.93&	3.24&	12.98&	4.93&	2.24&	2.86&	3.51&	1.91&	3.67\\
		& 0.0463&	0.0023&	0.0119&	0.0428&	0.0017&	0.0089&	0.0259&	\textbf{0.0009}&	0.0066&	0.6168&	0.1647&	0.4852\\		\hline
		
		Glass & 50.36&	56.29&	51.37&	52.91&	58.32&	54.78&	52.89&	57.35&	53.79	&	52.88&	\textbf{58.80}&	57.83\\
		214 $\times$ 9 $\times$ 6		& 4.28&	3.89&	2.16&	3.87&	3.50&	4.80&	2.98&	3.78&	4.92&	4.37&	3.26&	3.40\\
		& 0.04&	0.0018&	0.0187&	0.35&	0.0016&	0.0163&	0.0229&	\textbf{0.0011}&	0.0116&	0.6497&	0.2083&	0.4245
		\\
		\hline
		
		PB 	& 90.55&	87.81&	87.81&	88.09&	89.92&	90.44&	88.09&	90.54&	90.44&	87.55&	93.09&	\textbf{93.13}\\
		5473 $\times$ 10 $\times$ 5	&	0.81&	3.17&	3.17&	3.55&	1.51&	1.90&	3.55&	1.59&	1.65&	2.89&	0.88&	1.40\\
		&	2.01754&	0.0162&	0.0257&	1.9726&	0.0113&	0.0176&	1.3161&	\textbf{0.008}&	0.013&	563.64&	77.1921&	109.3996
		\\  
		\hline
		
		MF 	& 82.50&	85.25&	90.35&	82.25&	84.75&	75.60&	83.35&	84.40&	84.70&	\textbf{97.60}&	96.35&	96.25\\
		2000 $\times$ 649 $\times$	10 &	2.80&	4.65&	1.97&	1.25&	4.51&	6.36&	4.67&	4.83&	2.79&	0.87&	1.92&	2.45\\
		&	121.6527&	24.8009&	44.864&	86.8096&	20.8418&	36.8624&	70.8630&	\textbf{15.1547}&	28.2157&	520.0735&	15.1251	& 10.2126\\
	
		\hline
		
		OD & 88.25&	90.94&	90.64&	89.54&	92.43&	90.23&	89.84&	91.68&	\textbf{92.46}&	88.25&	90.94&	90.64\\
		5620 $\times$ 64 $\times$ 10	&	0.81&	0.47&	1.70&	1.65&	0.88&	1.45&	2.31&	0.52&	1.06&	1.03&	0.47&	1.70\\
		&	4.75&	0.0543&	0.1659&	3.89&	0.0412&	0.0973&	2.8513&	\textbf{0.0323}&	0.0680&	1263.4512&	0.0326&	5.2645
		\\
		\hline \hline
		 
Avg Acc	& 82.76&	85.14&	85.41&	83.23&	85.83&	84.97&	83.53&	86.59&	86.78&	85.29&	89.46& \textbf{89.73}\\
Avg SD	&	3.58&	3.98&	3.77&	4.00&	3.98&	5.38&	4.26&	3.12&	3.44&	3.64&	2.77&	3.03\\
Avg Time	&	14.19&	2.49&	4.51&	10.15&	2.09&	3.70&	8.45&\textbf{	1.52}&	2.83&	236.22&	9.35&	12.74\\ \hline
		              
	\end{tabular}
	\caption{Comparison of NHCAs with Linear Classifiers}
	\label{tab:ucilin}  
\end{table*}

For the nonlinear implementation, RBF kernel is used and kernel parameters are appropriately chosen through grid search method. The comparison results of nonlinear classifiers, on UCI datasets, are listed in Table \ref{tab:ucinl}. The table shows mean of accuracy over 5-folds and standard deviation, as well as average training time of the classifiers. The table demonstrates that the accuracy of the nonlinear classifier is better than that of the linear ones. Table \ref{tab:ucinl} shows that the classification results of TDS-TWSVM are best among all the algorithms, over ten datasets and mean accuracy is $92.91\%$. W-L-T for TWSVM and GEPSVM-based classifiers, considering classification accuracy, are 5-5-0 and 5-5-0, for classification accuracy. This shows that TWSVM and GEPSVM-based classifiers have comparable performance. Also, W-L-T for OAA, BT and TDS are 2-8-0, 2-7-1 and 5-4-1, which demonstrates that TDS excels other two approaches by bagging maximum wins, in terms of classification accuracy.

\begin{table*}[t]
	\centering

	\begin{tabular}{|l|c|c|c|c|c|c|c|c|c|c|c|c|}
		\hline
		\multirow{5}{*}{ } &
		\multicolumn{3}{c|}{IGEPSVM } &
		\multicolumn{3}{c|}{GEPSVM } &
		\multicolumn{3}{c|}{Reg GEPSVM} &
		\multicolumn{3}{c|}{TWSVM} \\
		\hline
		& OAA & BT & TDS & OAA & BT & TDS & OAA & BT & TDS & OAA & BT & TDS \\
		\hline
		
		& Acc  & Acc  & Acc  & Acc& Acc  & Acc& Acc  & Acc& Acc  & Acc& Acc  & Acc \\
	DATA SETS	& SD & SD & SD & SD & SD & SD & SD & SD & SD & SD & SD & SD \\
		\hline
		\hline
		Iris & 92.00&	94.67&	94.67&	89.33&	96.67&	96.67&	93.33&	96.00&	\textbf{98.00}&	94.00&	96.67&	97.33\\
&		6.91&	2.74&	1.83&	7.23&	2.36&	2.36&	2.36&	2.79&	1.82&	4.34&	2.35&	1.49
		 \\
		\hline
		
		Seeds & 90.86&	89.52&	90.48&	89.90&	90.86&	90.48&	92.86&	93.33&	\textbf{94.29}&	93.33&	94.28&	93.80\\
	&	3.22&	3.61&	3.76&	2.11&	3.22&	2.63&	3.37&	4.26&	2.13&	3.10&	3.61&	4.32
		  \\
		
		\hline
		Derm & 81.78&	95.82&	94.30&	82.77&	93.53&	95.05&	84.66&	95.05&	95.83&	92.80&	\textbf{96.96}&	96.59\\
	&	4.24&	3.14&	3.26&	6.17&	3.70&	2.56&	4.71&	2.56&	2.06&	4.67&	1.67&	2.45
		 \\
		\hline
		
		Wine & 84.52&	90.48&	96.59&	80.39&	85.86&	83.38&	93.76&	97.75&	98.86&	98.32&	98.88&	\textbf{99.43}\\
	&	3.11&	2.91&	3.73&	2.71&	4.02&	10.00&	5.51&	2.38&	2.56&	2.49&	1.52&	1.27
		 \\
		\hline
		
		Zoo & 82.56&	95.05&	86.10&	87.45&	90.05&	88.05&	89.05&	90.05&	91.05&	96.04&	\textbf{97.04}&	\textbf{97.04}\\
	&	5.50&	3.54&	9.67&	4.32&	2.12&	5.77&	5.54&	5.06&	4.24&	4.17&	2.69&	2.69
		 \\
		
		\hline
		
		Ecoli & 80.32&	84.02&	85.05&	\textbf{85.64}&	82.15&	85.26&	77.38&	84.52&	85.26&	76.76&	82.88&	87.47\\
	&	3.61&	3.46&	14.76&	3.01&	2.31&	12.45&	4.19&	1.25&	3.26&	3.99&	1.91&	3.19
		 \\
		\hline
		
		Glass & 65.23&	68.52&	69.23&	65.12&	66.48&	69.66&	64.32&	\textbf{71.23}&	70.12&	62.68&	70.54&	69.17\\
	&	3.62&	4.32&	4.56&	3.21&	3.11&	2.95&	2.01&	3.14&	3.12&	3.67&	4.90&	4.95
		 \\
		\hline
		
		PB 	& 92.02& 92.38& 92.33&	92.12&	92.56&	92.56&	93.49&	94.62&	\textbf{96.04}&	94.66&	92.89&	92.85\\
	&	0.63&	0.66&	0.78&	1.34&	0.98&	0.98&	1.71&	1.92&	0.76&	0.54&	3.64&	0.33
		 \\  
		\hline
		
		MF 	& 80.55&	84.65&	90.40&	87.20&	82.10&	83.00&	87.50&	83.25&	87.35&	98.20&	82.85&	\textbf{96.75}\\
	&	4.09&	4.94&	1.77&	5.41&	6.58&	3.64&	6.36&	5.48&	3.15&	0.54&	3.10&	0.85
		\\
		\hline
		
		OD & 80.55&	84.65&	90.40&	87.20&	82.10&	83.00&	87.50&	83.25&	87.35&	\textbf{98.20}&	82.85&	96.75\\
	&	4.09&	4.94&	1.77&	5.41&	6.58&	3.64&	6.36&	5.48&	3.15&	0.54&	3.10&	0.85
		 \\
		\hline \hline
		
Avg Acc	&	84.64&	88.93&	89.64&	85.48&	87.65&	87.93&	87.05&	90.23&	91.50&	89.97&	90.39&	\textbf{92.91}\\
Avg SD	&	3.62&	3.09&	4.50&	3.68&	2.94&	4.46&	3.71&	3.02&	2.43&	2.88&	2.59&	2.22\\
		\hline
		
	\end{tabular}
	\caption{Comparison of NHCAs with Nonlinear Classifiers}
		\label{tab:ucinl}  
\end{table*}

\section{Conclusions}
\label{sec:con}
In this paper, we have presented a comparative study of nonparallel hyperplanes classification algorithms (NHCAs) in multicategory framework. For this work, we have extended Generalized eigenvalue proximal SVM (GEPSVM), Regularized GEPSVM (RegGEPSVM), Improved GEPSVM (IGEPSVM) and Twin SVM (TWSVM) in multicategory scenario, using One-Against-All (OAA), binary tree-based (BT) and ternary decision structure (TDS) approaches. The experiments are conducted with ten benchmark UCI datasets. It is observed that TWSVM achieves higher classification accuracy as compared to GEPSVM-based classifiers, but TWSVM is computationally less efficient than GEPSVM-based classifiers. The use of TWSVM is recommended when the number of features are very high, as for UCI Multiple Features dataset with 649 features; for such datasets, GEPSVM-based classifiers do not perform well. It is also ascertained that GEPSVM-based classifiers performs better than TWSVM, with large datasets, in terms of learning time. The tree-based multicategory approaches are more efficient than OAA, regarding classification accuracy as well as learning and testing time. TDS requires $\lceil log_3K \rceil$ comparisons for evaluating test data as compared to $\lceil log_2K \rceil$ comparisons required by BT and $K$ comparisons required by OAA approaches. Thus, TDS requires minimum testing time. The experimental results show that TDS-TWSVM outperforms other methods in terms of classification accuracy and BT-RegGEPSVM takes the minimum time for building the classifier. This work can be extended by exploring other NHCAs \cite{ding} with different approaches for multicategory classification \cite{hsu2002comparison}.

\section*{Acknowledgment}
We would like to take this opportunity to thank Dr.Suresh Chandra, for his constant encouragement throughout the preparation of the manuscript.

\bibliography{Biblio}

\end{document}